# Unconstrained influence diagrams

Finn V. Jensen and Marta Vomlelová

Department of Computer Science

Aalborg University, Denmark

{fvj, marta}@cs.auc.dk

## Abstract

We extend the language of influence diagrams to cope with decision scenarios where the order of decisions and observations is not determined. As the ordering of decisions is dependent on the evidence, a *step-strategy* of such a scenario is a sequence of dependent choices of the next action. A *strategy* is a step-strategy together with selection functions for decision actions. The structure of a step-strategy can be represented as a DAG with nodes labeled with action variables. We introduce the concept of GS-DAG: a DAG incorporating an optimal step-strategy for any instantiation. We give a method for constructing GS-DAGs, and we show how to use a GS-DAG for determining an optimal strategy. Finally we discuss how analysis of relevant past can be used to reduce the size of the GS-DAG.

## 1 Introduction

The beautiful princess in the kingdom Lovania has a wooer. It is rather convenient for the king as he considers retirement. Furthermore, in case he starts a war with the neighbor king, he needs a good general. As customary, the king shall confront the wooer with three tasks. One of the tasks shall be either to kill a unicorn or a dragon. Another task will be to spend a night in the royal tomb or in the haunted castle tower. The third type of task is to swim across the river or to climb the highest mountain in the kingdom.

The king can decide to retire or to start a war at any time. However, he cannot start a war after retirement, and he cannot give his daughter to the wooer before he has been confronted with all three tasks. As the sequence of decisions is not settled, the king's decision problem cannot be represented through an influence diagram (Howard and Matheson 1981; Shachter 1986). The problem may be represented through a decision tree, but as decision trees grow exponentially with the number of variables, it is not an attractive representation language. You may also think of the problem as an asymmetric decision scenario and use one of the representation languages for that (Covaliu and Oliver 1995; Shenoy 2000; Nielsen and Jensen 2000). However, these languages cannot be used here without introducing artificial variables and creating a very large representation.

Another framework is a partial influence diagram (Nielsen and Jensen 1999). Partial influence diagrams (PIDs) were introduced to represent decision scenarios were the linear temporal order of decisions was relaxed, but where the maximal expected utility is the same no matter how the order is extended to a linear order (so-called *well defined* PIDs). The PID for the king's problem is not well defined, and to solve it, we need to find a sequencing of the decisions maximizing the expected utility. As the optimal choice of next step may depend in the decisions and observations in the past, an optimal solution to the king's problem may have a proper network structure.

In this paper we extend the language of IDs with extra features and we shall call the graphs *unconstrained influence diagrams* (UIDs). We present syntax and semantics of the language, and we give a method for solving UIDs.

## 2 The representation language

To describe the decision scenario for the king, we have to specify which variables are observable. To do so, we let observable chance variables be doubly circled. This is done in Figure 1.

An *unconstrained influence diagram* (UID) is a DAG

over decision variables (rectangular shaped), chance variables (circular shaped), and utility variables (diamond shaped). Furthermore, utility variables have no children. There are two types of chance variables, *observables* (doubly circled) and *non-observables* (singly circled).

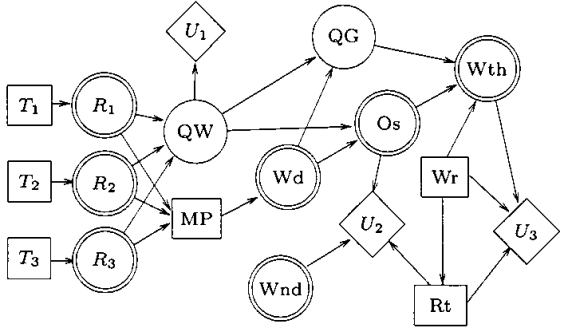

Figure 1: An unconstrained influence diagram for the king's decision scenario. Abbreviations used: $T_i$ (task i), $R_i$ (result of task i), $Wnd$ (wooer's noble descent), $QW$ (quality of wooer), $QG$ (quality of general), $MP$ (marry princess), $Wd$ (wedding), $Os$ (offspring), $Rt$ (retire), $Wth$ (wealth), $Wr$ (war).

The quantitative specification required is similar to the specification for influence diagrams: conditional probabilities and utility functions. We add the convention that each decision variable D has a cost. If this cost only depends on D, it is not represented graphically, and the cost function is attached to D. We say that an UID is *instantiated* when the structure has been extended with the required quantitative specifications.

The semantics of an UID is similar to the semantics of IDs. A link into a decision variable represents informational precedence; a link into a chance variable represents causal influence; a link into a utility variable represents functional dependence. We assume *no-forgetting*: at each point of the decision process the decision maker knows all previous decisions and observations.

We add a semantic clarification, which is not necessary for influence diagrams. As the order of observations and decisions is not determined by the structure, it might seem that a descendant of a decision node may be observed prior to the decision. This will have no meaning, and therefore descendants of a decision node should be regarded as non-existing until the decision is made. If you have the option of observing a variable before and after an ancestral decision, this should be modeled through two different variables.

On the other hand, an observable can be observed when all its antecedent decision variables have been decided upon. In that case we say that the observable

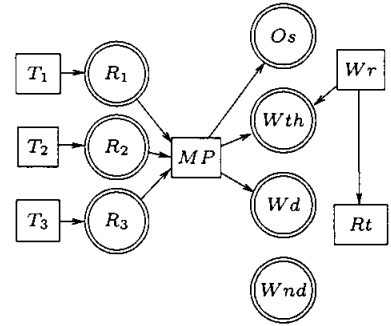

Figure 2: The partial temporal order induced by Figure 1.

is *free*, and we *release* an observable when the last decision in its ancestral set is taken.

The structural specification yields a partial temporal order. The temporal order for Figure 1 is shown in Figure 2.

If the structure is extended to a linear ordering we get an influence diagram. Such an extension is called an *admissible order*. The problem addressed in Nielsen and Jensen (1999) is whether all admissible orderings yield the same optimal strategy.

When dealing with UIDs, the concept of *strategy* is more complex then is the case for IDs. In principle we look for a set of rules telling us what to do given the current information, where "what to do" is to choose the next action as well as choosing a decision option if the next action is a decision. Notice that the choice of next action may be dependent on the specific information from the past.

**Notation** Let $\Gamma$ be an unconstrained influence diagram. The set of decision variables is denoted $\mathcal{D}_\Gamma$, the set of observables is denoted $\mathcal{O}_\Gamma$. Let $\mathcal{X} \subseteq \mathcal{D}_\Gamma \cup \mathcal{O}_\Gamma$ be a set of variables; $sp(\mathcal{X})$ denotes the set of configurations over $\mathcal{X}$ (ignoring order). The partial temporal order induced by $\Gamma$ is denoted $\prec_\Gamma$. When obvious from the context we avoid the subscript.

**Definition 1** *Let $\Gamma$ be an UID.*

*An S-DAG is a directed acyclic graph G. The nodes are labeled with variables from $\mathcal{D}_\Gamma \cup \mathcal{O}_\Gamma$ such that each maximal directed path in G represents an admissible ordering of $\mathcal{D}_\Gamma \cup \mathcal{O}_\Gamma$. (Figure 3 gives an example of an S-DAG for the king's problem).*

*Let N be a node in an S-DAG. The* history *of N (denoted hst(N)) is the union of labels of N and its ancestors. The union of labels of N's children is denoted ch(N). A* step-policy *for N is a function $\sigma : sp(hst(N)) \rightarrow ch(N)$.*

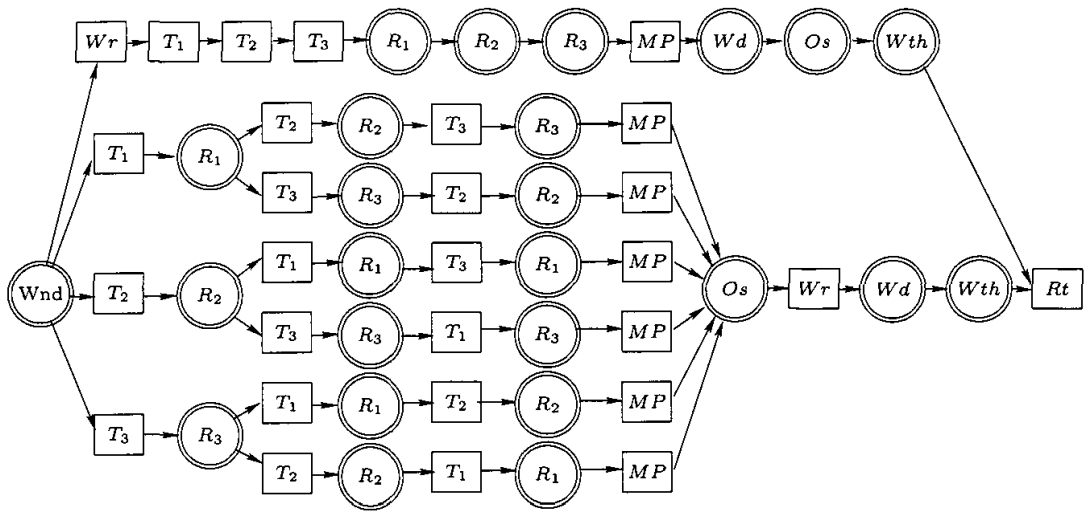

Figure 3: An S-DAG for the king's problem.

A step-strategy *for* $\Gamma$ *is a couple* $(\Sigma, \mathcal{S})$, *where* $\Sigma$ *is an S-DAG for* $\Gamma$ *and* $\mathcal{S}$ *is a set of step-policies, one for each node in* $\Sigma$.

A policy *for N is an extension of a step-policy, such that whenever the step-policy yields a decision variable D, then the policy yields a state of D. A* strategy *for $\Gamma$ is an S-DAG together with a policy for each node.*

We now need to define the concept of *expected utility* (*EU*) of a strategy for UID. As a precise definition is a bit complex we shell not give it here. Instead, notice that any strategy $S$ for an UID can be folded out to a *strategy tree*: following the policies in $S$ we construct a tree where all root-leaf paths represent an admissible ordering. The expected utility of a strategy tree is defined as for decision trees, and the expected utility of a strategy is the expected utility of the corresponding strategy tree.

A *solution* to an UID is a strategy of maximal *EU*. Such a strategy is called *optimal*.

## 3  Normal form S-DAGs

We wish to construct an S-DAG which is guaranteed to contain an S-DAG for an optimal strategy. Our concern is to construct it as small as possible. To reduce the S-DAG we use the following two observations

1. The expected utility can never increase by delaying an observation[1].

   So, we need not have any path with a decision variable placed before a free observation.

2. As two maximizations (summations) over finite variables are commutable, a sequence of variables of the same type can be commuted without changing the *EU*. So, a sequence of consecutive variables of the same type can be characterized as a set rather than a sequence.

Due to 2. we let the labels of the nodes be sets rather than single variables. When it causes no confusion we will not distinguish between a node and its label, and when the label consists of one variable, we avoid talking about it in set terms. Using 1. we restrict ourselves to S-DAGs in *normal form*: each parent of a node labeled with an observable $V$ contains a decision node $D$ such that $D \prec V$. Furthermore, nodes with identical history have the same children.

Figure 4 gives a normal form S-DAG for the king's problem.

As the labels of nodes in a normal form S-DAG are sets of variables of the same type, we classify them as decision nodes and observation nodes. Note that an observation node has only decision nodes as children, and a decision node $D$ may have as children either a single observation node $O$ or a set of decision nodes. The *decision children* of $D$ are in the first case the children of $O$ and in the latter case the children of $D$.

**Definition 2** *The* skeleton *of a normal form S-DAG G is a DAG over G's decision nodes. There is an edge from D to D' if and only if D' is a decision child of D in G.*

Figure 5 shows the skeleton of the normal form S-DAG in Figure 4. Note that the normal form S-DAG can easily be reconstructed from its skeleton.

---
[1] Observations are cost free.

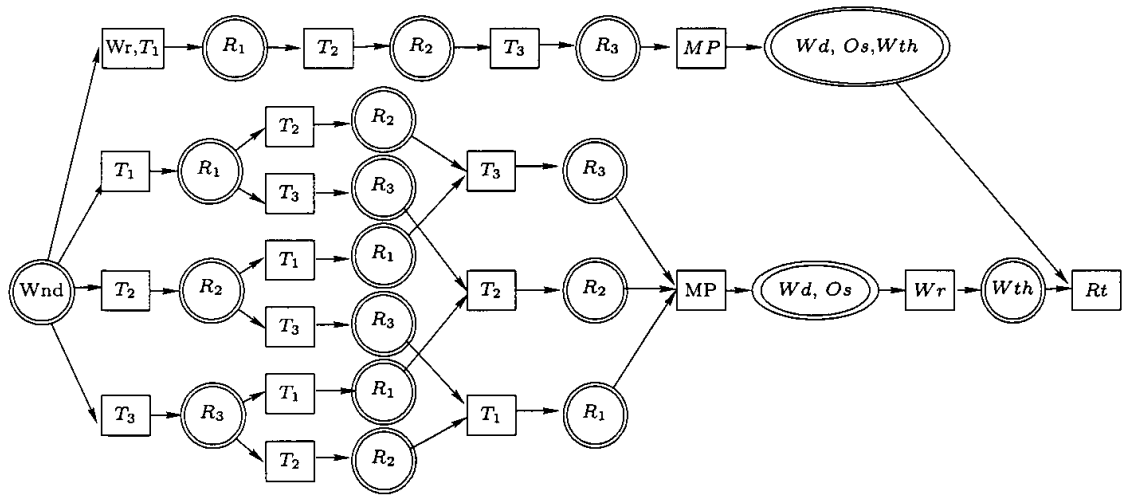

Figure 4: A normal form S-DAG for the king's problem.

We aim at constructing a *GS-DAG*: an S-DAG which is guaranteed to include an S-DAG for an optimal strategy. Due to 1. and 2. we only need to search among normal form S-DAGs.

## 4 Construction of GS-DAGs

We wish to construct a GS-DAG as small as possible. In this section we present an algorithm exploiting some simple rules reducing the size. In Section 6 we shall present other reduction rules.

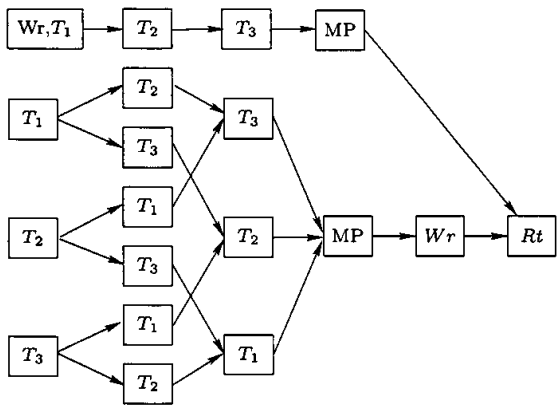

Figure 5: The skeleton of the S-DAG from Figure 4.

Instead of presenting the general algorithm, we shall show how it works in the king's problem. We construct the skeleton of a GS-DAG, and the construction works in reverse temporal order. That is, we start off considering which decision can be taken last.

From the partial order in Figure 2 we see that only $Rt$ and $MP$ can be the last decision. Consider the situation where $MP$ is last. Then the observations $\{Os, Wd, Wt\}$ must follow this decision, and $Rt$ comes before $MP$. (See Figure 6)

If the child of $Rt$ is an observation, this observation does not require $Rt$, and $Rt$ can be moved to the right (1. above). The same holds if the child of $Rt$ is a decision. So eventually, $Rt$ is the last variable.

In the next step we have to consider $Wr$ and $MP$. For the same reason as above, $MP$ cannot come after $Wr$ ($Wr$ can be commuted with everything except $Wth$ and $Rt$). We get the partial skeleton in Figure 7.

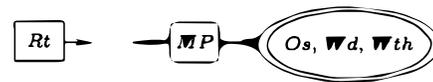

Figure 6: The sequencing if $MP$ is the last decision.

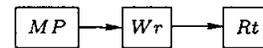

Figure 7: A partial skeleton during the construction of a GS-DAG for the king's problem.

Now, we can choose among the $T_i$s, and the skeleton is branched. Figure 8 shows the skeleton after two branchings.

When incorporating the last $T_i$ in the skeleton we notice that some of nodes will have identical history. Therefore, these nodes can be identified. We end up with the skeleton in Figure 10, and with the GS-DAG in Figure 9).

**Notation** Let $G$ be a partially constructed skeleton for the UID $\Gamma$ and let $T$ be a top node of $G$.

The *future* of $T$ is the set of the decision nodes together with the released observables on the maximal directed paths of $G$ starting in $T$ (and including $T$).

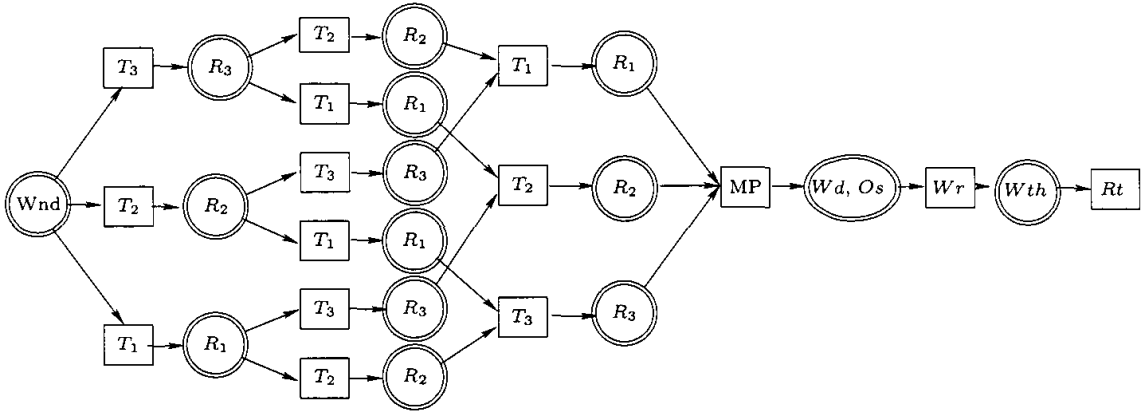

Figure 9: The GS-DAG constructed.

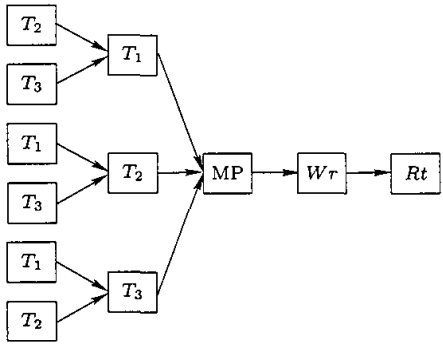

Figure 8: The partial skeleton after two $T_i$-branchings.

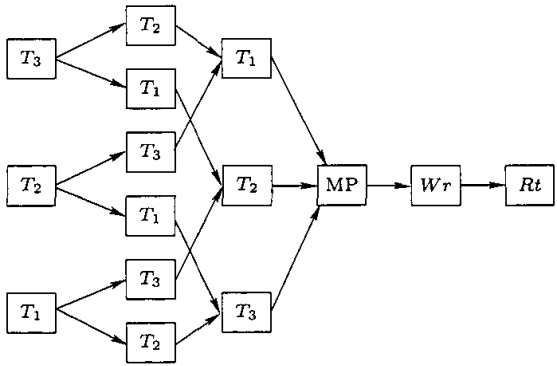

Figure 10: The final skeleton.

The decision variable $D$ in $\Gamma \setminus future(T)$ is *co-free for $T$*, if all its descendant decision nodes in $\Gamma$ are members of the future of $T$.

**Proposition 4.1** *Let $D_1$ and $D_2$ be co-free for the top node $T$. Let $\Omega(D_i)$ denote the set of observations $O$ for which $D_i \prec O$, and which are not members of the future of $T$. If $\Omega(D_1) \subset \Omega(D_2)$, then $D_2$ shall not be selected as a parent of $T$. If $\Omega(D_1) = \Omega(D_2)$ we construct a common parent labeled with $D_1$ and $D_2$.*

**Proof outline.** Assume that $D_2$ is selected. Then, in the eventual GS-DAG $D_1$ is an antecedent of $D_2$. As no observation on any path from $D_1$ to $D_2$ has $D_1$ as an ancestor in the UID, $D_1$ can be commuted with all variables on that path (including $D_2$). √

Algorithm 1 gives the pseudo code for the construction algorithm. In the pseudo code, every node $N$ in the GS-DAG under construction is uniquely defined by a pair of sets $[label(N), future(N) \setminus label(N)]$. If a node with the same label and future is already generated, it is re-used instead of generated again.

**Algorithm 1 (Generating GS-skeleton)**

$G \leftarrow$ empty graph; **process** $\leftarrow$ empty list;
**free** $\leftarrow$ all decisions without observable descendants
       (possibly ∅);
$G \leftarrow$ add node [**free**, ∅]
**process** $\leftarrow$ add node [**free**, ∅]
**while process** not empty
   **pNode** $\leftarrow$ take first node from **process**
   **if** exists a decision not in $future$(**pNode**)
      **parents** $\leftarrow$ find_parents(**pNode**)
      **for** every **parent** ∈ **parents**
         **if** node **parent** does not exist in $G$
            create it and add it in **process**

```
      add link from parent to pNode
   endfor
endwhile

function find_parents(pNode)
   for every D ∉ future(pNode)
      de(D) ← desc(D)\ future(pNode)
   endfor
   remove all sets de(D) such that
      there is a D_1: de(D_1) ⊂ de(D)
   create sets eq(D) that contains
      all decisions D_1: de(D_1) = de(D)
   return the list
      of nodes [eq(D), future(pNode)]
```

From the reasoning above we conclude this section with

**Proposition 4.2** *Algorithm 1 yields a skeleton of the GS-DAG.*

## 5 Solving a GS-DAG

A GS-DAG is solved in almost the same manner as for influence diagrams (Shachter 1986; Shenoy 1992; Jensen et al. 1994).

We eliminate variables in reverse order. When a branching point is met, the elimination is branched out, and when several paths meet, the probability potentials are the same, and the utility potentials are unified through maximization. To illustrate the method we use the UID in Figure 11 with GS-DAG in Figure 12.

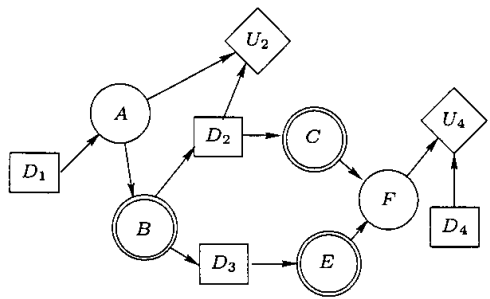

Figure 11: An UID.

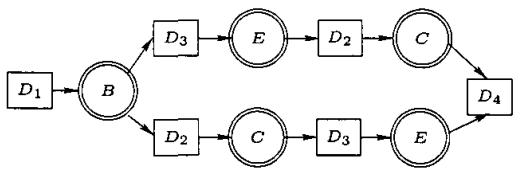

Figure 12: A GS-DAG for the UID in Figure 11.

Due to personal biases we illustrate the method using the lazy propagation method (Madsen and Jensen 1999). We start off with the two sets

$$\Phi = \{P(A|D_1), P(B|A), P(C|D_2), P(E|D_3), P(F|C,E)\},$$
$$\Psi = \{U_1(D_1), U_2(A, D_2), U_3(D_3), U_4(F, D_4)\}.$$

First the non-observables are eliminated. A chance variable $X$ is eliminated in $\Phi$ by multiplying all potentials with $X$ in the domain to get $\phi'_X$ and (sum)marginalizing $X$ out of $\phi'_X$ to get $\phi_X$. To eliminate $X$ out of $\Psi$ we take the sum of all utility potentials with $X$ in the domain, multiply this with $\phi'_X$ and marginalize $X$ out. The result is divided by $\phi_X$.

When $A$ and $F$ are eliminated we get the sets

$$\Phi' = \{P(B|D_1), P(C|D_2), P(E|D_3)\},$$
$$\Psi' = \{U_1(D_1), U'_2(B, D_1, D_2), U_3(D_3), U'_4(C, E, D_4)\},$$

where

$$P(B|D_1) = \Sigma_A P(A|D_1) P(B|A);$$
$$U'_2(B, D_1, D_2) = \frac{1}{P(B|D_1)} \Sigma_A P(A|D_1) P(B|A) U_2(A, D_2);$$
$$U'_4(C, E, D_4) = \Sigma_F P(F|C, E) U_4(F, D_4).$$

Note that $\Sigma_F P(F|C, E)$ is the neutral potential.

We eliminate a decision variable $D$ by taking the sum of the utility potentials with $D$ in the domain and max-marginalizing $D$ out. At the same time, a choice function is determined. When a decision variable is to be eliminated it cannot be in the domain of any probability potential. When $D_4$ has been eliminated we have

$$\Psi^4 = \{U_1(D_1), U'_2(B, D_1, D_2), U_3(D_3), U''_4(C, E)\},$$

where $U''_4(C, E) = max_{D_4} U'_4(C, E, D_4)$,
and $\delta_{D_4}(C, E) = argmax_{D_4} U'_4(C, E, D_4)$.

Next we branch, and produce one set of potentials after elimination of $C$ and another set after eliminating $E$.

$$\Phi^C = \{P(B|D_1), P(E|D_3)\}$$
$$\Psi^C = \{U_1(D_1), U'_2(B, D_1, D_2), U_3(D_3), U^C_4(E, D_2)\},$$

where $U_4^C(E, D_2) = \Sigma_C P(C|D_2)U_4''(C, E)$. And

$$\Phi^E = \{P(B|D_1), P(C|D_2)\}$$
$$\Psi^E = \{U_1(D_1), U_2'(B, D_1, D_2), U_3(D_3), U_4^E(C, D_3)\},$$

where $U_4^E(C, D_3) = \Sigma_E P(E|D_3)U_4''(C, E)$.

When eventually $D_3$ has been eliminated in the $C$-branch, and $D_2$ is eliminated in the $E$-branch, we have the two potential sets.

$$\Phi^{Ce} = \{P(B|D_1)\},$$
$$\Psi^{Ce} = \{U_1(D_1), U^C(B, D_1)\};$$
$$\Phi^{Ee} = \{P(B|D_1)\},$$
$$\Psi^{Ee} = \{U_1(D_1), U^E(B, D_1)\}.$$

It is no coincidence that the two probability potential sets are identical. They are both the result of sum-marginalizing the same set of variables from the same set of potentials. As sum-marginalizations can be commuted, the two branches must give the same result. Before marginalizing B we unify the utility potentials sets by taking the max:

$$\Psi^{Ee} = \{U_1(D_1), max(U^C(B, D_1), U^E(B, D_1))\}$$

The step function

$$\sigma(b, d_1) = \begin{cases} D_3 & \text{if } U^C(b, d_1) \geq U^E(b, d_1); \\ D_2 & \text{otherwise.} \end{cases}$$

The book keeping of potentials is rather easy to handle. After an elimination, create a set of new potentials (or scripts for computing them); and for the other potentials, just keep a pointer.

Our implementation was able to solve the king's problem in 30 seconds. Solving the worst-case[2] problem with nine decisions took 8 minutes. For the worst-case problem with ten decisions the GS-DAG was created, but the evaluation of it ran out of memory.

---

[2] The worst case, with no structural constraints in the UID, forces maximal branching of the GS-DAG, i.e. $\approx 2^{|D|}$ nodes. In the worst case, we need to store a full utility table of size $\approx 2^{|D|}$ in each node, which for our system caused memory overflow for ten decisions.

## 6 Exploiting irrelevance

As stated already above, the algorithm presented in Section 4 provides a sufficiently "fat" S-DAG. However, as the solution of a GS-DAG requires heavy table operations, we wish to work with a GS-DAG as slim as possible. As the table operations are resource demanding, we can afford to spend rather much time on graph algorithms for trimming the GS-DAG.

**Definition 3** *Let $D$ be a decision variable in an influence diagram, and let $past(D)$ be the decisions and observations performed. A variable $A \in past(D)$ is* structurally relevant *for $D$ if there is an instantiation such that the optimal policy for $D$ is a proper function of $A$. $A$ is* irrelevant *if it is not structurally relevant.*

Various methods for determining structural relevance in influence diagrams have been developed (Shachter 1998; Nielsen and Jensen 1999; Lauritzen and Nilsson 2001). It holds for IDs as well as for UIDs that the sequencing of the past does not matter. However, the sequencing of the future does matter. We shall return to this in section 7.

Let $T$ be a top node in a partially constructed skeleton. Let $D_1$ and $D_2$ be candidates for being parents of $T$, and let $O_i$ be the observations released by $D_i$. If, for example, information on $\{D_1, O_1\}$ is not relevant for $D_2$, then we need not include $D_2$ as a parent of $T$, and the decision $D_2$ can be pushed forward (see Figure 13).

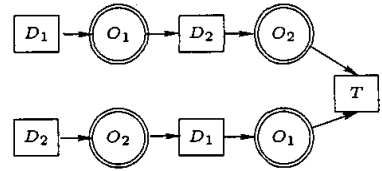

Figure 13: If $\{D_1, O_1\}$ is irrelevant for $D_2$, then the upper branch cannot yield a higher $EU$ than the lower one can.

This means that we can extend the algorithm such that at each point of adding parents to a top node we analyze for irrelevance. For any candidate $D$, if another candidate $D'$ together with its released observations are irrelevant for $D$, then $D$ is not included as a parent. If all candidates are mutually irrelevant we take any of them.

## 7 Relevance analysis for UIDs

Relevance analysis in UIDs can be performed in much the same way as in Nielsen and Jensen (1999) for PIDs.

**Definition 4 (Lauritzen and Nilsson (2001))**

*Let $D$ be the last decision in an influence diagram. A variable $A \in past(D)$ is requisite for $D$ if there is a utility node $U$ such that given $past(D) \setminus \{A\}$ there is an active path[3] from $A$ to $U$ and $U$ is a descendant of $D$.*

**Proposition 7.1** *If $A$ is not requisite for $D$, then $A$ is irrelevant for $D$.*

An influence diagram is analyzed for irrelevance by starting with the last decision $D$. Let $req(D)$ be the set of requisite variables in $past(D)$. Then $D$ is substituted by a chance variable with $req(D)$ as parents, and in this way we recursively analyze the decision variables in reverse temporal order.

In the case of UIDs, consider a top node $T$ (see Figure 14). As the future for $T$ may be performed in different orders, we have to analyze them all.

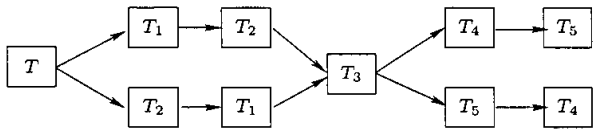

Figure 14: A partial skeleton where the top node $T$ has four different extensions. There are four different IDs to analyze for each candidate.

Let $req_1(D), \ldots, req_m(D)$ be the sets of requisite variables determined for a candidate $D$. Then the final set is the union of the $req_i$s.

It is up to further research to establish ways of accumulating information in the process such that we only need to visit $T$'s neighbors for the analysis. Also, it should be investigated whether it might be more efficient to use the solution method in Section 5 directly to establish relevance for the instantiated UID.

## 8 Conclusions

The system was offered to the king, and he decided to retire immediately and hand over the royal decisions to the marvelous system with us operating it.

## Acknowledgments

We wish to thank the decision support systems group at Aalborg University for fruitful discussions. In particular, we thank Thomas D. Nielsen for valuable feedback. We also thank the authors of the Elvira[4] system for providing a basis for our implementation.

---

[3]$U$ is not d-separated from $A$ given $past(D) \setminus \{A\}$.
[4]http://leo.ugr.es/~elvira/